\documentclass[sigconf]{acmart}

\usepackage{graphicx}
\usepackage{subcaption}

\AtBeginDocument{%
  \providecommand\BibTeX{{%
    \normalfont B\kern-0.5em{\scshape i\kern-0.25em b}\kern-0.8em\TeX}}}

\setcopyright{acmcopyright}
\setcopyright{rightsretained}
\copyrightyear{2021}
\acmYear{2021}
\acmDOI{}
\acmISBN{} 
\copyrightyear{2021}
\acmYear{2021}

\acmConference[KDD Humanitarian Mapping Workshop '21]{ACM SIGKDD Conference on Knowledge Discovery and Data Mining
}{August 14-18, 2021}{Virtual Conference}
\acmBooktitle{KDD Humanitarian Mapping Wokshop '21: ACM SIGKDD Conference on Knowledge Discovery and Data Mining
August 14-18, 2021; Virtual Conference}

\begin{document}

\title[Mapping illegal waste dumping sites with neural-network classification of satellite imagery]{Mapping illegal waste dumping sites with neural-network classification of satellite imagery}


\author{Maria Roberta Devesa}
\affiliation{%
  \institution{Dymaxion Labs}
  \city{Buenos Aires}
  \country{Argentina}}
\email{ro.devesa@dymaxionlabs.com}

\author{Antonio Vazquez Brust}
\affiliation{%
  \institution{Fundación Bunge y Born}
  \city{Buenos Aires}
  \country{Argentina}}
\email{avazquez.fellow@fundacionbyb.org}

\renewcommand{\shortauthors}{Devesa and Vazquez Brust}

\begin{abstract}

Public health and habitat quality are crucial goals of urban planning. In recent years, the severe social and environmental impact of illegal waste dumping sites has made them one of the most serious problems faced by cities in the Global South, in a context of scarce information available for decision making. To help identify the location of dumping sites and track their evolution over time we adopt a data-driven model from the machine learning domain, analyzing satellite images. This allows us to take advantage of the increasing availability of geo-spatial open-data, high-resolution satellite imagery, and open source tools to train machine learning algorithms with a small set of known waste dumping sites in Buenos Aires, and then predict the location of other sites over vast areas at high speed and low cost. This case study shows the results of a collaboration between Dymaxion Labs and Fundación Bunge y Born to harness this technique in order to create a comprehensive map of potential locations of illegal waste dumping sites in the region.

\end{abstract}





\acmISBN{}
\acmDOI{}

\maketitle

\section{Introduction}
Waste management is a a critical utility, essential for urban life. It is also a significant challenge for municipal governments all over the world. Often rated among the top priorities faced by cities in developing countries, it can also be taken for granted, missing from national or international political agendas \cite{wilson_waste_2015}.  The problem is aggravated in developing regions such as Latin America where the rapidly increasing production of waste, coupled with burdened municipal budgets and scarce national attention, result in poor waste management \cite{hoornweg_managing_2007}. Dumping sites, areas where intentionally and illegally abandoned waste accumulates, are a physical manifestation of this inadequacy. Dumping sites generate severe environmental degradation that disproportionately affects the most vulnerable population \cite{pellow_politics_2004}, specially those who -without access to better options- occupy land near or even inside dumping sites. This close proximity of human settlement to informal waste disposal areas has been associated to severe health risks such as respiratory disease \cite{ibrahim_impacts_2021} and heavy metal poisoning \cite{cittadino_heavy_2020}. Due to their unplanned and illicit nature, both the appearance and the growth of dumping sites are hard to monitor and can evolve unnoticed by governments. This lack of visibility is part of a larger issue, as precise data on the many aspects of waste management is severely lacking all over the world. In 2015 an international team tasked by the UN to measure the societal and environmental impact of poor waste management found that "availability and reliability of waste and resource data is dire, and urgently needs attention" \cite{wilson_waste_2015}. As with other pressing developing world issues where a chronic absence of timely information hinders decision-making, the exploration of remote sensing approaches has the potential to yield useful and frequently updated results at a fraction of the cost associated with on-the-ground data collection.

\section{Related Work}
As far back as 1998 Algarni and Elsadiq \cite{ali_algarni_mapping_1998} reported their experience identifying waste disposal sites in the City of Riyadh (Saudi Arabia) and its surroundings, applying digital processing techniques to SPOT imagery to facilitate visual detection by experts using GIS software. GIS-aided classification methods, taking advantage of spatial modeling features from widely available GUI based software, continue to be developed. Gill et al. \cite{gill_detection_2019} used Landsat thermal data to detect unauthorised landfills in Kuwait by their heat generation, measuring ground surface temperature and isolating suspiciously hot areas; Vambol et al. \cite{vambol_nature_2019} experimented with a GIS-based workflow to detect illegal waste dumps in the Kharkiv region (Ukraine) by analyzing the pixel brightness values of high resolution satellite imagery obtained via Google Earth. On the other hand, as expected given the fast development of general-purpose classification algorithms during the last decades, recent experiences on dumping detection aim to scale beyond traditional spatial modeling by relying on machine learning techniques. Akinina et al. \cite{akinina_methods_2017} compared the performance of three different algorithms applied to illegal dump detection using Landsat data for the Ryazan Region in central Russia: Parzen window, AdaBoost and neural network classification, with the latter yielding the best results both in accuracy and speed. Skogsmo \cite{skogsmo_scalable_2020} proposed a support vector machine (SVM) approach for large-scale analysis of Sentinel data, along with a case study of automated detection of waste dumps in Kampala, Uganda. Due to our use of machine learning rather than a GIS approach, our work most closely resembles the last two examples. Compared to those, a key difference is our implementation of a deep learning model instead of alternative classification algorithms like SVM or decision trees.

\section{Materials}

\subsection{Study area}
This study focuses on the Buenos Aires Province in Argentina. The province has an area of 307,571 km²,  although most of its population is concentrated on its largest conurbation, the Buenos Aires Metropolitan Area (occupying 3,830 km²). During the last decades urban solid waste production grew rapidly in the region, overflowing the capacity of publicly run waste disposal centers and propitiating the multiplication of illegal open air waste dumping sites. The environmental and societal impact associated with the proliferation of dumping sites made them one of the most conflictive patterns of land use change in the region, and a significant risk for public health \cite{cittadino_atlas_2012}. This situation makes Buenos Aires a relevant area for experimentation with large-scale, high-frequency detection of waste dumping sites. 

\subsection{Data Sources}

Satellite images used in this research are from the Sentinel-2 Earth observation mission, retrieved via Google Earth Engine. The image resolution used is 10 meters, and they were captured in the 2021-02-01 - 2021-02-28 period. After bench-marking results with different combinations, the best performance was obtained using the three color bands R-G-B, the near-infrared (IR), and the 2 near-infrared short wave bands (SWIR-1 and SWIR-2).  

Two data sets with georeferenced locations of known illegal waste dumps were available: A survey from the \textit{Autoridad de Cuenca Matanza Riachuelo} (a public agency tasked with improving the environmental quality of the region's main river basin), and a report from the Buenos Aires Province ombudsman. Together they provide 81 annotations for open air waste dumps, which were used to train the supervised machine learning algorithm presented in this study.

\section{Method}

\begin{figure*}[h!]
 \includegraphics[scale=0.3]{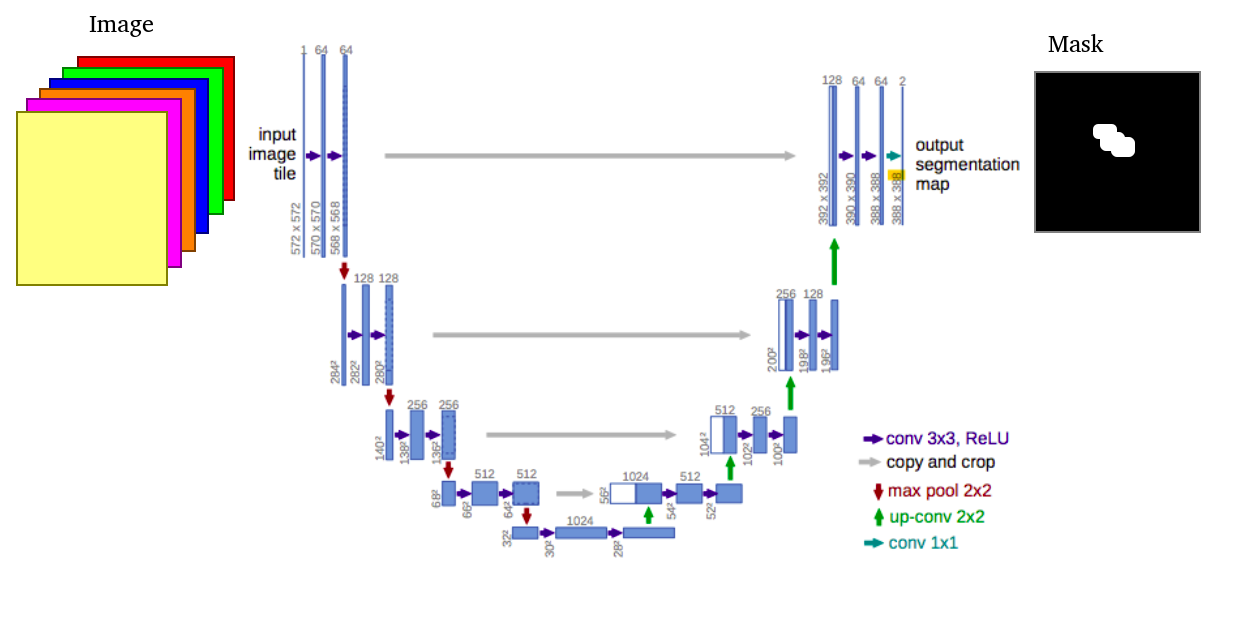}
 \caption{Diagram of the training process}
 \label{fig:d}
\end{figure*}

This study implements a Convolutional Neural Network (CNN) model by a U-Net architecture developed for a segmentation process that identifies the characteristic patterns of dumping sites. The output of the model will be an image of the same size as the input, in which each pixel is the probability of belonging to a waste dump. This means that the model will find not only their position but also their boundaries. This feature is of great importance for planning applications, as it allows measuring the growth of sites over time, as well as their potential intersection with other land uses (watercourses, inhabited areas, etc). The model needs an image set and annotations for the training process. Each image has an associated binary mask that delimits the object of interest. The image can have many bands (besides just RGB) so additional information can be included as a band.  Figure~\ref{fig:d} shows a diagram of the training process. This study uses 6-band images: three color bands (RGB), a near-infrared (NIR) band, a short wave infrared (SWIR-1) band, and the normalized difference of the two SWIR. Previous studies of methane emission detection inspired the use of combined SWIR-1 and SWIR-2 bands, on the assumption that open-air waste dumps may emit a significant amount of methane, particularly from organic waste.
NIR and SWIR bands are used because they are more sensitive to green areas. As large dumping sites are usually located in rural areas, these bands may help to classify the background better. Table~\ref{fig:metric} shows the improvement when these bands are included.

The following subsection details each step of the process.

\subsection{Pre-processing}
\begin{figure*}[h!]
 \includegraphics[scale=0.8]{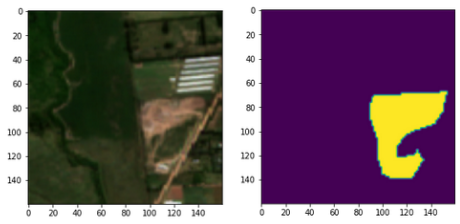}
 \caption{A example of an image (left) and its respective mask (right) used to train the model}
 \label{fig:masks}
\end{figure*}

The preprocessing stage aims to create the dataset used to train the model and predict with it. A combination of tools from GDAL and pysatproc were used to create image sets with the chosen bands and masks for each one. In addition, the pysatproc tool sets the image size for the model and allows to perform a sliding window over an annotation to increase the number of images available for training. As an example, the Figure~\ref{fig:masks} shows an image with its mask.

\subsection{The model}
As it was briefly mentioned before, the U-Net is a type of CNN model. It was initially created for biomedical image segmentation \cite{unet_bib} although it also works for the segmentation of natural images.
The U-Net is constructed by multiple down-sampling and up-sampling layers, which are connected by concatenating layers. This way it combines the high precision activation nodes with nodes of the previous layer with better resolution.

\subsection{The weighted binary cross entropy method}
Under imbalanced data, the model tends to recognize background more than objects of interest, thus learning to predict background terrain more than objects. To deal with this condition the model uses weighted binary cross-entropy \cite{loss}. It allows us to consider different weight values for each class (or one class, in this case), to force the loss function to focus more on the cases that are less frequent on the dataset.
\subsection{Post-processing}
The prediction process produces a set of images with multiple bands as classes, where each one has the probability of finding the object of interest in the image, pixel by pixel. Our case only considers one class, so the output consists of one band images.

To remove cases with low probability a filter process is implemented over all predicted results, applying per-pixel probability threshold. Additionally, the resulting images were polygonized to remove predicted objects based on surface area, as the smaller ones are more likely to be false positives. The threshold was set at 100 m², as waste dumping sites are usually larger than that.

\section{Results}

A U-net model was performed to segment pixels showing dumping sites from satellite images. The U-net architecture was built with four down-sampling layers and four up-sampling layers, with six channels in the first convolutional layer, and a 3×3 kernel size in all of the model's convolutional layers.

The dataset contains 1917 images with a size of 100x100 pixels per 6-band of depth images and their respective masks.
It was then divided into training, validation, and test. The first two were used to perform the model and the last one to obtain the metrics over untouched data. A split ratio of 0.1 for testing, and a split ratio of 0.7 for training and validation. 
The metric analyzed in this study is the Intersection over Union (IoU), which is mainly used in applications related to object detection and object segmentation. It is essentially a method to quantify the percent overlap between the target mask and our prediction output, measuring the number of pixels common between the target and prediction masks divided by the total number of pixels present across both masks.
The model was set with batch 16 and epoch 30, but after 18 epochs the model reaches the plateau; with a 0.0458 loss and a mean IoU of 0.675.

Table-\ref{fig:metric} shows the comparison of the model with different approaches. The loss and the IoU values reached over the test sample are compared for a different combination of bands on the input images. It shows that the optimal performance is obtained using RGB-NIR-SWIR-NDSW
\begin{table}[h!]
\begin{center}

 \begin{tabular}{|c | c | c |} 
 \hline
 Image information & Loss & IoU \\ [0.5ex] 
 \hline\hline
 RGB & 0.0613 & 0.593 \\ 
 \hline
 RGB-NIR & 0.0712 & 0.660 \\
 \hline
 RGB-NIR-SWIR & 0.0652 & 0.650 \\
 \hline
 RGB-NIR-SWIR-NDSW & 0.0458 & 0.675 \\
 \hline

\end{tabular}
\caption{Table of the metrics for each approach.}
\label{fig:metric}
 
\end{center}
\end{table}

\begin{figure*}[h!]
 \includegraphics[scale=0.7]{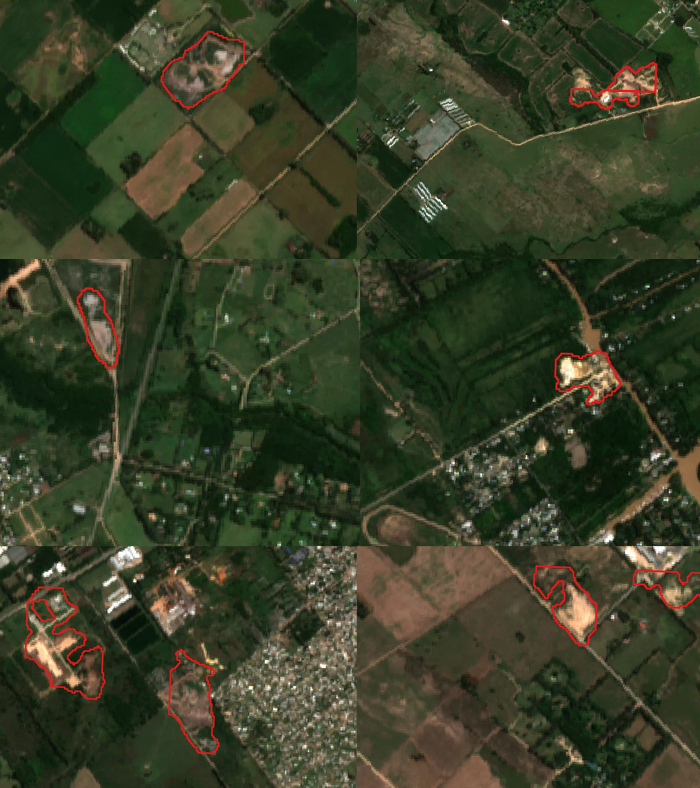}
 \caption{Results for different locations around Buenos Aires city. The red line shows the area detected by the model as a potential illegal waste dumping site.}
 \label{fig:pred}
\end{figure*}

As an example Figure~\ref{fig:pred} shows some model result examples, obtained after processing images from 594 urban centers across the Buenos Aires province. The images were originally captured between 2021-02-01 and 2021-02-28.

Some of the main limitations found after running the prediction process over the different zones of Buenos Aires were that despite that the model detected several waste dumps, also other places like home constructions on the initial state were incorrectly tagged. 

In addition, small waste dumps (size less than 1 hectare) are hard to detect due to the image resolution. And in some areas may be hard to get images without clouds.

\section{Conclusions and future work}
This research proposes a supervised machine learning method to detect illegal waste dumping sites using satellite imagery. It consists of a U-Net architecture based on CNN  for classification and segmentation and weighted cross-entropy. The model's performance in predicting IoU is 0.6304 and it reaches a loss of 0.0413.  

To get further insights into model performance, future work can include images from earlier years and additional annotations to train the model and predict with it.

From a replication perspective, working with Sentinel- 2 satellite images is critical not only for the current study but also for future iterations because of their public availability and ease of acquisition. Free access to 10 meter resolution satellite imagery paired with state-of-the-art open source machine learning algorithms enable the fast analysis of extensive areas at a minimal cost. 

To facilitate monitoring growth of illegal dumping sites, we are currently packaging our methodology and algorithms as cloud-based programming notebook, and hosting hands-on training sessions aimed at municipal personnel.  

\bibliographystyle{ACM-Reference-Format}
\bibliography{biblio.bib}

\section{About the authors}

Maria Roberta Devesa is PhD. in high energy physics. Currently, works in Computer Vision and Machine Learning since more than 4 years and has more than 7 years of experience in data analysis in different fields.

Antonio Vazquez Brust holds a MS degree in Urban Informatics. His area of interest is the application of computational analysis to study urban phenomena. He collaborates with NGOs and international agencies to better understand and govern territorial processes.

\end{document}